\def\year{2022}\relax
\documentclass[letterpaper]{article} 
\usepackage{aaai22}  
\usepackage{times}  
\usepackage{helvet}  
\usepackage{courier}  
\usepackage[hyphens]{url}  
\usepackage{graphicx} 
\usepackage{natbib}  
\usepackage{caption} 
\DeclareCaptionStyle{ruled}{labelfont=normalfont,labelsep=colon,strut=off} 
\usepackage{algorithm}
\usepackage{algorithmic}
\usepackage{physics}
\usepackage{amsfonts}
\usepackage{booktabs}
\usepackage{placeins}
\usepackage{mathtools}
\usepackage{graphics}
\usepackage{graphicx}
\usepackage{tabularx}
\usepackage{multirow}
\usepackage{arydshln}
\usepackage{amsmath}
\usepackage{newfloat}
\usepackage{listings}
\floatstyle{ruled}
\newfloat{listing}{tb}{lst}{}
\floatname{listing}{Listing}
\title{Model2Detector:Widening the Information Bottleneck for Out-of-Distribution \\
Detection using a Handful of Gradient Steps}
\author {
    Sumedh A Sontakke\thanks{Work done as a research intern at Nokia Bell Labs. Corresponding Author: ssontakk@usc.edu},\textsuperscript{\rm 1}
    Buvaneswari Ramanan, \textsuperscript{\rm 2}
    Laurent Itti \textsuperscript{\rm 1}
    Thomas Woo \textsuperscript{\rm 2}
}
\affiliations {
    \textsuperscript{\rm 1} University of Southern California\\
    \textsuperscript{\rm 2} Nokia Bell Labs, Murray Hill, NJ\\
    
}

\usepackage{bibentry}

\begin{document}

\maketitle

\begin{abstract}
\label{sec:abstract}
Out-of-distribution detection is an important capability that has long eluded vanilla neural networks. Deep Neural networks (DNNs) tend to generate over-confident predictions when presented with inputs that are significantly out-of-distribution (OOD).  
This can be dangerous when employing machine learning systems in the wild as detecting attacks can thus be difficult.  
Recent advances inference-time out-of-distribution detection help mitigate some of these problems. However, existing methods can be restrictive as they are often computationally expensive. Additionally, these methods require training of a downstream detector model which learns to detect OOD inputs from in-distribution ones. This, therefore, adds latency during inference. Here, we offer an information theoretic perspective on why neural networks are inherently incapable of OOD detection. We attempt to mitigate these flaws by converting a trained model into a an OOD detector using a handful of steps of gradient descent. Our work can be employed as a post-processing method whereby an inference-time ML system can convert a trained model into an OOD detector. Experimentally, we show how our method consistently outperforms the state-of-the-art in detection accuracy on popular image datasets while also reducing computational complexity.
\end{abstract}

\section{Introduction}
\begin{figure*}[htp!]
    \centering
    \includegraphics[width=\textwidth]{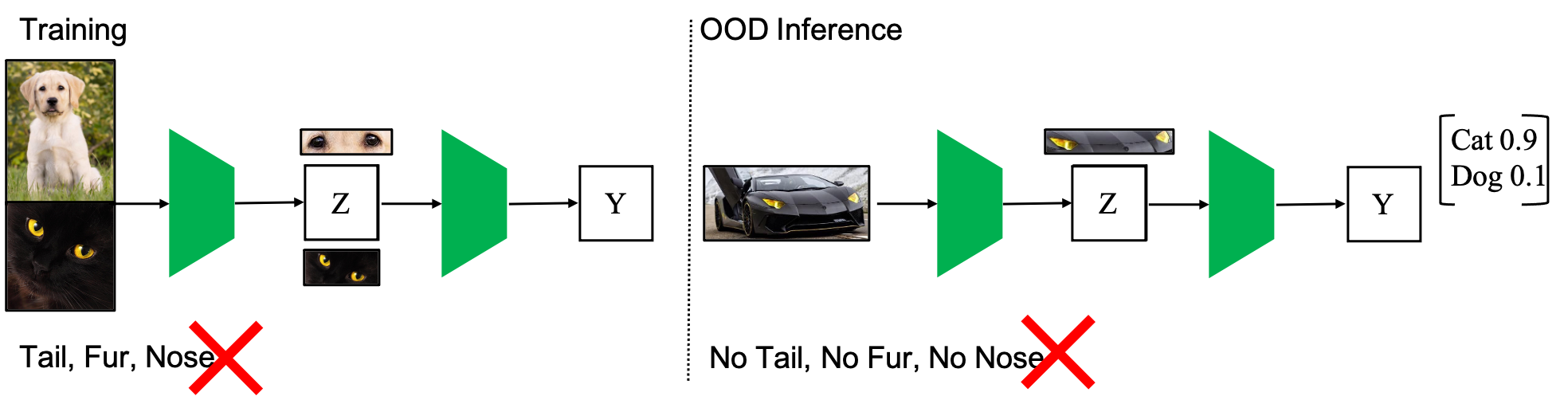}
    \caption{\textbf{Shared features across classes are discarded.} During training a DNN learns to differentiate between images by encoding discriminative features in its internal representation while discarding shared features across classes. These shared features are useful for out-of-distribution detection. Above, features like tail, fur, etc., are discarded while discriminative features like the shape and color of the eyes are encoded. Thus when presented with an OOD input, the DNN produces overconfident incorrect outputs. }
    \label{fig:oodInference}
\end{figure*}
Recent advances in the Information Bottleneck (IB) theory have helped improve the ability to analyze and understand how DNNs make decisions \citep{tishby2015deep}. The IB framework can be used to evaluate the optimality of the internal representation generated by a DNN. Further, it has been used to make claims about the properties of stochastic gradient descent (SGD) and the computational benefit of deep networks \citep{shwartz2017opening, saxe2019information}. Consider that $\mathbf{Y}$ is the ground-truth label, $\mathbf{X}$ are features at the input of the DNN, and $\mathbf{Z}$ is either a latent representation or the output of the DNN for the input $\mathbf{X}$. The IB functional as defined by \citet{tishby2000information} is
\begin{equation}
\label{eq:IB}
    I(\mathbf{X};\mathbf{Z}) - \beta I(\mathbf{Y};\mathbf{Z})
\end{equation}
where $I(A;B)$ is defined as the mutual information between random variables $\mathbf{A}$ and $\mathbf{B}$ and $\beta > 1$ is a parameter balancing some trade-off. Equation (\ref{eq:IB}) is subsequently minimized during stochastic gradient descent. 

Intuitively, the objective above implies that the DNN behaves as a kind of compressing/gating mechanism, regulating the flow of information through its layers. At the input layer, it receives $\mathbf{X}$ with a Kolmogorov Complexity $C_f(\mathbf{X})$. Subsequently, during training, the network attempts to learn an internal representation $\mathbf{Z}$ such that $\mathbf{Z}$ can be used to predict $\mathbf{Y}$ (second term in Equation (\ref{eq:IB})) while throwing away all information available at the input layer from $\mathbf{X}$ not useful in predicting $\mathbf{Y}$ (first term in Equation (\ref{eq:IB})). Thus, $C_f(\mathbf{X}) > C_f(\mathbf{Z})$.

While the IB objective is useful in building high performing predictive models, the very nature of the objective prevents the classifier from being able to reject OOD inputs. We motivate this problem with an example. Consider a binary classifier DNN trained on a dataset of cats and dogs. Equation (\ref{eq:IB}) implies that the DNN will only encode information about $\mathbf{X}$ in $\mathbf{Z}$ which allows it to predict $\mathbf{Y}$ while discarding all information in $\mathbf{X}$ not relevant to the prediction task. Thus, the classifer will only extract features about the cats and dogs that help it differentiate them.

Thus, as shown in Figure (\ref{fig:oodInference}), the network will encode information about discriminative features between cats and dogs (for instance the shape and color of the eyes) while discarding the shared of features between the classes (for instance, the fact that both cats and dogs have fur, a tail, etc.). 
On the contrary, such shared features are useful in detecting whether an input to the DNN is in distribution or not. For instance, the fact that cats and dogs have tails and fur while cars have neither are useful in being able to detect that an image of a car is OOD to such a cat-vs-dog DNN classifier. Thus it's no surprise that vanilla DNNs generate overconfident predictions for OOD data; they "see" patterns where there are none.

In our work, we attempt to recover these features shared across classes in order to aid OOD detection. To achieve this, we "widen" the information bottleneck using a retraining objective optimized on a trained DNN. The objective helps increase $I(\mathbf{X};\mathbf{Z})$ allowing the the internal representation of the network to contain information about non-discriminatory features of the input data. We show that these features can be used to detect OOD samples.

The contributions of our work are detailed below:
\begin{itemize}
    \item We offer an \textbf{Information Theoretic perspective} on out-of-distribution detection and discuss how DNNs discard shared features across classes during training. 
    \item We propose a post-processing method to ensure that such shared features are encoded. These features are subsequently used to train a simple linear model as a OOD detector. 
    \item We demonstrate the efficacy of our approach on popular vision and language datasets. We also demonstrate how the resultant detector can be applied in parallel with a classifier thereby reducing the latency in high volume inference systems. 
\end{itemize}

\section{Related Work}
Out-of-Distribution Detection with Deep Networks. \citet{hendrycks2016baseline} demonstrate that a deep, pre-trained classifier has a lower maximum softmax probability on anomalous examples than in-distribution examples, so a classifier can conveniently double as a consistently useful outof-distribution detector. Building on this work, \citet{devries2018learning} attach an auxiliary branch onto a pre-trained classifier and derive a new OOD score from this branch. \citet{liang2017enhancing} present a method which can improve performance of OOD detectors that use a softmax distribution. In particular, they make the maximum softmax probability more discriminative between anomalies and
in-distribution examples by pre-processing input data with adversarial perturbations \citep{goodfellow2014explaining}. 
\citet{lee2017training} train a classifier concurrently with a GAN (\citet{radford2015unsupervised}; Goodfellow et al., 2014), and the classifier is trained to have lower confidence on GAN samples. For each testing distribution of anomalies, they tune the classifier and GAN using samples from that out-distribution. Unlike \citet{liang2017enhancing}; \citet{lee2017training}, Many other works (de Vries et al., 2016; Subramanya et al., 2017; Malinin and Gales, 2018; Bevandic et al., 2018) also encourage the model to have lower confidence on anomalous examples. Recently, \citet{liu2018can} provide theoretical guarantees for detecting out-of-distribution examples under the assumption that a suitably powerful
anomaly detector is available.
\section{Methods}
\begin{figure*}
    \centering
    \includegraphics[width=\textwidth]{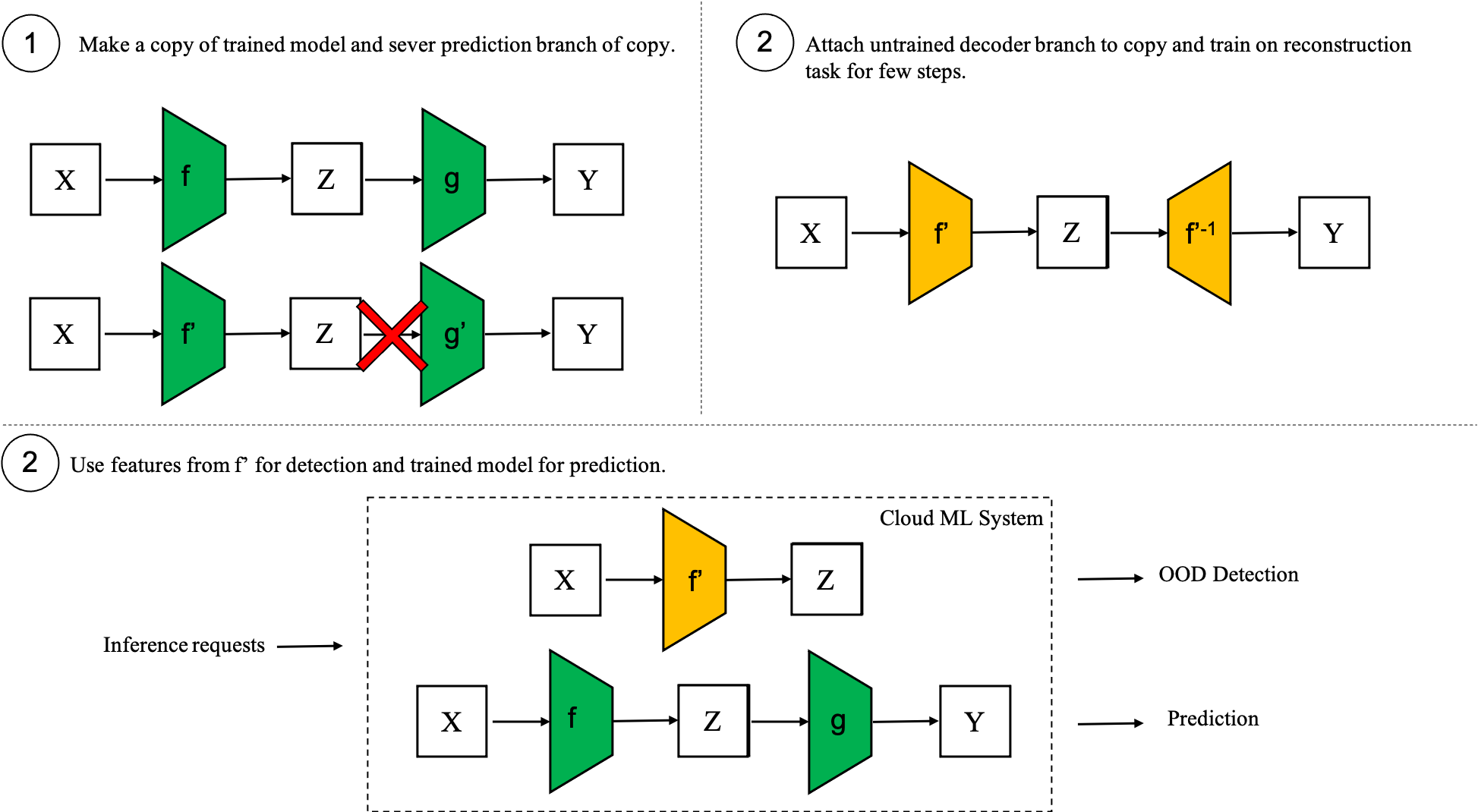}
    \caption{\textbf{Method Overview.} A copy of the model is made. The copy is then severed at the prediction branch and an untrained decoder is connected to the internal layer. Subsequently, the Autoencoder ensemble is trained for a few steps of the gradient on a reconstruction objective. Finally, the retrained encoder is then placed in parallel to the trained branch during inference. The reconstruction loss post-processing ensures that the retrained encoder encodes more than just the discriminative features resulting easier OOD Detection. }
    \label{fig:overview}
\end{figure*}

\subsection{Setup}
We consider the scenario of deployment of a model in-the-wild. A trained model is hosted in a cloud-based system providing inference-as-a-service. Users send inference requests to the system and the systems serves them.
We assume that system has access to the model parameters and a small subset of the training data using which the model was trained. 

We consider the task of deciding whether or not a sample is from a learned distribution called $D_{in}$.
Samples from $D_{in}$ are called “in-distribution,” and otherwise are said to be “out-of-distribution” (OOD) or samples from $D_{out}$. In real applications, it may be difficult to know the distribution
of outliers one will encounter in advance. Thus, we consider the realistic setting where $D_{out}$ is unknown. We do, however, assume that some OOD data is available to train the detector.  
\subsection{Encoding General Features}
The trained model is first duplicated. Next, the prediction branch of the copy is severed. Subsequently, an untrained decoder is connected to the embedding layer of the copy. The coupled model is then trained using a few steps of the stochastic gradient descent on the reconstruction objective:
\begin{equation}
    \mathcal{L}_{recon} = CE(\mathbf{y}, \mathbf{\hat{y}})
\end{equation}
The objective is trained for a handful of gradient descent steps using:
\begin{equation}
    \mathbf{w}' = \mathbf{w} - \eta \nabla \mathcal{L}_{recon}
\end{equation}
This retrained DNN is used as an OOD detector. We conduct ablation studies varying the number of gradient steps that achieves optimal performance. We also vary which layer is used to generate the input to the decoder in our experiments. 
\subsection{OOD Detection}
We assume that the features produced by the detector module are Gaussian conditioned on the class, i.e., $P(f(\mathbf{x}|\mathbf{y}=c))=\mathcal{N}(\mathbf{\mu_{c}, \mathbf{\Sigma}})$ the parameters of which are learnt empirically from data.
\begin{multline}
    \mathbf{\hat{\mu_c}}=\frac{1}{N_c}\sum_{i:y_i=c}f(\mathbf{x}_i),\\ \mathbf{\Sigma}=\frac{1}{N}\sum_{c}\sum_{i:y_i=c}(f(\mathbf{x}_i)-\mathbf{\hat{\mu_c}})(f(\mathbf{x}_i)-\mathbf{\hat{\mu_c}})^T
\end{multline}
A Mahalanobis distance-based confidence score is then calculated using:
\begin{equation}
    C(\mathbf{x}) = \max_c -(f(\mathbf{x}_i)-\mathbf{\hat{\mu_c}})^T \mathbf{\Sigma}^{-1}(f(\mathbf{x}_i)-\mathbf{\hat{\mu_c}})
\end{equation}
Intuitively, the Mahalanobis distance between the mean of the closest class and the generated features, $f(\mathbf{x})$ for a given input $\mathbf{x}$ is used as a measure of confidence for predicting whether the $\mathbf{x}$ is OOD or otherwise.
\section{Experiments}
In this section, we demonstrate the effectiveness of the proposed method using deep convolutional neural networks such as DenseNet \citep{huang2017densely} and ResNet \citep{he2016deep} on various vision datasets: CIFAR \citep{krizhevsky2009learning},
SVHN \citep{netzer2011reading}, ImageNet \citep{deng2009imagenet} and LSUN \citep{yu2015lsun}. 
\subsection{Setup.} For the problem of detecting out-of-distribution (OOD) samples, we train DenseNet with 100 layers and ResNet with 34 layers for classifying CIFAR-10, CIFAR-100 and SVHN datasets. The dataset used in training is the in-distribution (positive) dataset and the others are considered as OOD (negative). We only use test datasets for evaluation. In addition, the TinyImageNet (i.e., subset of ImageNet dataset) and LSUN datasets are also tested as OOD. For evaluation, we use a thresholdbased detector which measures some confidence score of the test sample, and then classifies the test sample as in-distribution if the confidence score is above some threshold. We measure the following metrics: the area under the receiver operating characteristic curve (AUROC), and the detection accuracy. For comparison, we consider the baseline method \citep{hendrycks2016baseline}, which defines a confidence score as a maximum value of the posterior distribution, and the state-of-the-art ODIN \citep{liang2017enhancing}, which defines the confidence score as a maximum value of the processed posterior distribution.

\subsection{Results}
Table \ref{table:Main} validates the contributions of our suggested techniques under the comparison with the baseline method and ODIN. We measure
the detection performance using ResNet trained on CIFAR-10, when SVHN dataset is used as OOD.
We incrementally apply our techniques to see the stepwise improvement by each component. One
can note that our method significantly outperforms the baseline method without feature ensembles and input pre-processing. This implies that our method can characterize the OOD samples very effectively compared to the posterior distribution. By utilizing the feature ensemble and input preprocessing, the detection performance are further improved compared to that of ODIN. Table \ref{table:AUROC_resnet} reports the detection performance with the state-of-the-art Mahalanobis. We find that our method significantly outperforms \citet{lee2018simple} for the same number of features. Table \ref{table:ablation} shows that the efficacy of our approach over similar training time Autoencoder style architectures.

\begin{table*}[]
\begin{tabularx}{\textwidth}{c|c|X|X|X|X}
\multicolumn{1}{l}{Pretraining} & OOD              & 1 Layer, 5 step & 1 Layer, 10 step & 1 Layer, 100 step & 1 Layer Mahalanobis \\
\hline
\multirow{3}{*}{cifar10}        & svhn             & 66.39/63.43           & 70.56/67.61           & \textbf{72.35/68.18}    & 65.8/62.98            \\
                                & imagenet\_resize & 91.75/84.6           & \textbf{92.54/85.52}   & 85.46/78.02             & 91.53/84.26           \\
                                & lsun\_resize     & 93.05/85.92           & \textbf{93.9/87.05}    & 85.9/78.56              & 93.15/86.03           \\
\hdashline
\multirow{3}{*}{cifar100}       & svhn             & 59.89/61.44           & \textbf{66.06/66.09}   & 62.15/62.44             & 59.92/61.13           \\
                                & imagenet\_resize & 90.3/82.73            & \textbf{90.66/83.21}   & 87.69/79.96             & 90.09/82.6           \\
                                & lsun\_resize     & 93.22/86.49           & \textbf{93.43/86.71}   & 89.21/82.03             & 93.36/86.62           \\
\hdashline
\multirow{3}{*}{svhn}           & cifar10          & 84.23/77.03          & 83.73/76.78            & \textbf{84.58/76.77}    & 84.66/77.37           \\
                                & imagenet\_resize & \textbf{95.82/89.54}  & 94.34/87.18            & 93.38/85.94             & 93.88/86.9           \\
                                & lsun\_resize     & \textbf{97.2/91.68}   & 95.37/88.43            & 93.68/86.34             & 95.27/88.59          
\end{tabularx}

\caption{\textbf{AUROC/Detection Accuracy compared to Mahalanobis \citep{lee2018simple}}. We train a resnet classifier on a variety of popular image datasets and test their efficacy on detecting out of distribution images from complementary datasets.}
\label{table:AUROC_resnet}
\end{table*}
\begin{table*}[]
\begin{tabularx}{\textwidth}{c|X|X|X|X}
\multicolumn{1}{l}{Pretraining} & OOD              & Tau-Softmax \tiny{\citep{hendrycks2016baseline}} & ODIN \tiny{\citep{liang2017enhancing}}       & Model2Detector \tiny{OURS}\\
\hline
\multirow{3}{*}{cifar10}        & svhn             & 89.9/85.1   & 96.7/ 91.1 & 91.32/84.77    \\
                                & imagenet\_resize & 91/85.1     & 94.0/86.5  & 98.37/94.4     \\
                                & lsun\_resize     & 91.0/ 85.3  & 94.1/86.7  & 99.04/95.97    \\
\hdashline
\multirow{3}{*}{cifar100}       & svhn             & 79.5/73.2   & 93.9/88    & 92.5/85.92     \\
                                & imagenet\_resize & 77.2/70.8   & 87.6/80.1  & 95.67/89.12    \\
                                & lsun\_resize     & 75.8/69.9   & 85.6/78.3  & 96.59/91.09    \\
\hdashline
\multirow{3}{*}{svhn}           & cifar10          & 92.9/90.0   & 92.1/89.4  & 96.59/90.47    \\
                                & imagenet\_resize & 93.5/90.4   & 92.0/89.4  & 99.41/97.54    \\
                                & lsun\_resize     & 91.6/89.0   & 89.4/87.2  & 99.48/        
\end{tabularx}
\caption{\textbf{AUROC/Detection Accuracy of Resnet Detector.} We train a resnet classifier on a variety of popular image datasets and test their efficacy on detecting out of distribution images from complementary datasets.}
\label{table:Main}
\end{table*}

\begin{table*}[]
\begin{tabularx}{\textwidth}{ll|r|r|r}
\multicolumn{1}{l}{Pretraining} & OOD              & 4 Layer 10 step & Vanilla AE 10 steps       & Best Performing\\
\multirow{3}{*}{cifar10}  & svhn             & 91.32                               & 55.61                                   & \textbf{72.35}                      \\
                          & imagenet\_resize & 98.37                               & 90.22                                   & \textbf{92.54}                      \\
                          & lsun\_resize     & 99.04                               & 91.27                                   & \textbf{93.9}                       \\

\hdashline
\multirow{3}{*}{cifar100} & svhn             & 92.5                                & 56.94                                   & \textbf{66.06}                      \\
                          & imagenet\_resize & 95.67                               & 86.1                                    & \textbf{90.66}                      \\
                          & lsun\_resize     & 96.59                               & 86.41                                   & \textbf{93.43}                      \\
\hdashline
\multirow{3}{*}{svhn}     & cifar10          & 96.59                               & 92.83                                   & \textbf{84.58}                      \\
                          & imagenet\_resize & 99.41                               & \textbf{99.07}                          & 95.82                               \\
                          & lsun\_resize     & 99.48                               & \textbf{99.3}                           & 97.2                               
\end{tabularx}
\caption{\textbf{Ablation on accuracy on the effect of pretraining.} We compare our approach to a vanilla untrained Autoencoder and find that our method outperforms it. This experiment confirms the utility of the pre-training.}
\label{table:ablation}
\end{table*}
\section{Conclusion}
In this work, we introduce \emph{-Model2Detector}, an inference-time post-processing method that can be used to convert classifiers into out-of-distribution detectors. We compare our methods against several state-of-the-art approaches and find that our approach offers lower computational complexity and higher accuracy.

\clearpage
\bibliography{aaai22}

\end{document}